# Creativity Has Left the Chat:
# The Price of Debiasing Language Models

Behnam Mohammadi[1]

Carnegie Mellon University

Tepper School of Business

Jun 8, 2024

## Abstract

Large Language Models (LLMs) have revolutionized natural language processing but can exhibit biases and may generate toxic content. While alignment techniques like Reinforcement Learning from Human Feedback (RLHF) reduce these issues, their impact on creativity, defined as syntactic and semantic diversity, remains unexplored. We investigate the unintended consequences of RLHF on the creativity of LLMs through three experiments focusing on the Llama-2 series. Our findings reveal that aligned models exhibit lower entropy in token predictions, form distinct clusters in the embedding space, and gravitate towards "attractor states", indicating limited output diversity. Our findings have significant implications for marketers who rely on LLMs for creative tasks such as copywriting, ad creation, and customer persona generation. The trade-off between consistency and creativity in aligned models should be carefully considered when selecting the appropriate model for a given application. We also discuss the importance of prompt engineering in harnessing the creative potential of base models.

## 1.      Introduction

Large Language Models (LLMs) have demonstrated remarkable capabilities in generating human-like text, with applications spanning various domains, including marketing. However, LLMs have also been shown to exhibit biases and generate toxic or inappropriate content (Bender et al., 2021; Gehman et al., 2020), prompting the development of techniques such as Reinforcement Learning from Human Feedback (RLHF) to *align* LLMs with human values and preferences (Ouyang et al., 2022; Stiennon et al., 2022), aiming to mitigate these issues.

While RLHF has proven effective in reducing biases and toxicity in LLMs, our work suggests that this alignment process may inadvertently lead to a reduction in the models'

---

[1] behnamm@cmu.edu

*creativity* and output diversity. In the context of this paper, we define "creativity" as the model's ability to generate outputs with high syntactic and semantic diversity. Syntactic diversity refers to the variety in the choice of words, sentence structures, and other linguistic elements, while semantic diversity pertains to the range of meanings, sentiments, and ideas expressed in the generated text[1].

The potential trade-off between safety and creativity is particularly relevant in the context of marketing, where generating diverse and engaging content is crucial for various applications, such as customer persona generation, ad creation, writing product description, and customer support. One specific application of LLMs in marketing is the generation of simulated customers or personas with diverse preferences and backgrounds. These personas can be used for various purposes, such as training bank associates to better communicate with actual customers or providing business school students with more engaging alternatives to traditional case studies.

When generating customer personas using LLMs, there are two primary approaches: generating multiple personas simultaneously[2] or generating them one at a time. While creating multiple personas simultaneously might seem more efficient, it has limitations due to the context size[3] constraints of LLMs and the causal attention mechanism in these models (Vaswani et al., 2017). The causal attention mechanism means that the distribution of the generated personas will not be independent, as each new persona would depend on the previous generations. Therefore, generating personas one at a time is a more suitable approach, as it solves both the context size and independence issues. However, when using this method with an aligned LLM, an unexpected challenge arises: the generated personas often exhibit striking similarities in their preferences and characteristics, lacking the desired heterogeneity. This lack of heterogeneity in the generated personas is problematic as it limits the ability to capture the diverse preferences and behaviors of real-world customers, potentially leading to less effective marketing strategies and suboptimal user experiences.

This observed lack of creativity in the outputs of aligned models led to the suspicion that the RLHF process itself might be the underlying cause. We investigate this problem by taking a foundational approach and examining the issue at both the semantic and syntactic levels. Our study comprises three experiments that aim to provide a comprehensive understanding of how the alignment process affects the diversity of LLM outputs.

---

[1] It is important to note that our use of the term "diversity" does not refer to the concept of diversity in the context of diversity, equity, and inclusion (DEI) or other similar domains, although one of our experiments does show that the aligned model exhibits reduced diversity in that sense as well.

[2] For example, asking the LLM to generate 100 personas delimited by new lines.

[3] Context size refers to the maximum number of tokens a transformer can handle simultaneously, encompassing both the input sequence and the generated output. For instance, a context size of 4,096 tokens, as seen in Meta's Llama-2 models, corresponds to approximately 6 pages of English text ("Context length in LLMs," 2023).

Experiment 1 serves as a concrete example of the impact of RLHF on creativity in a practical marketing context. We generate customer personas and their corresponding product reviews using both the base and aligned models, comparing the diversity of the generated attributes, such as names, demographics, and review content. The results reveal significant differences in the variety of outputs between the two models, with the aligned model exhibiting less diversity and more repetitive patterns.

Experiment 2 investigates the semantic diversity of the models' outputs by examining their ability to recite a historical fact about Grace Hopper[1] in various ways. The generated outputs are encoded into sentence embeddings and visualized using dimensionality reduction techniques. The results reveal that the aligned model's outputs form distinct clusters, suggesting that the model expresses the information in a limited number of ways. In contrast, the base model's embeddings are more scattered and spread out, indicating a higher level of semantic diversity in the generated outputs. These results are further supported by the cosine similarity analysis which shows the aligned model's outputs are more semantically similar to each other compared to the base model's outputs.

An intriguing property of the aligned model's generation clusters in Experiment 2 is that they exhibit behavior similar to *attractor states* in dynamical systems. We demonstrate this by intentionally perturbing the model's generation trajectory, effectively nudging it away from its usual output distribution. Surprisingly, the aligned model gracefully finds its way back to its own attractor state and in-distribution response. The presence of these attractor states in the aligned model's output space is a phenomenon related to the concept of *mode collapse* in reinforcement learning, where the model over-optimizes for certain outputs, limiting its exploration of alternative solutions. This behavior contrasts with the base model, which exhibits greater flexibility and adaptability in its outputs.

Experiment 3 delves into the syntactic diversity of the models by analyzing the entropy[2] of generated tokens and the probability distributions over the top predicted tokens at each step. The results show that the base model exhibits significantly higher average entropy, assigning more spread-out probabilities to different tokens, while the aligned model has a more skewed probability distribution, favoring certain tokens over others.

---

[1] The inventor of the COBOL programming language which is still heavily used by financial institutions and banks ("Grace Hopper," 2024).
[2] Entropy is a measure of uncertainty in a random variable.

The findings from these experiments suggest that the RLHF process, which aims to reduce biases and toxicity in LLMs, may transform them into more deterministic algorithms that lack the capacity to explore diverse sets of token trajectories, leading to reduced semantic and syntactic diversity in their outputs. In other words, aligned models exhibit higher *confidence* in their outputs, providing consistency and predictable behavior. However, this confidence comes at the cost of lowered creativity, as the models tend to stick to a limited set of outputs.

In marketing, this trade-off between consistency and creativity has far-reaching consequences. Applications such as copywriting, writing scripts for ad clips, and customer persona generation all require a high level of variation and diversity in the generated content. If an aligned model is used for these tasks, the resulting outputs may lack the necessary heterogeneity and novelty to effectively engage the target audience. Similarly, in the domain of *recommendation systems*, LLMs that lack diversity in their outputs may struggle to recommend a diverse set of products to users, potentially leading to suboptimal user experiences and reduced customer satisfaction.

It is important to note that base models, while more creative, are not directly usable in applications like chatbots. As a result, techniques such as *prompt engineering* (also known as prompt programming) (Sahoo et al., 2024) become even more crucial when working with base models. These techniques can help guide the models' outputs and make them more suitable for specific applications while still leveraging their creative potential. Contrary to the belief that prompt engineering may become obsolete, our findings suggest that these techniques will be more important than ever in harnessing the power of base models.

Consequently, the choice between base and aligned models should be carefully considered based on the specific requirements of the task at hand. For applications where creativity is paramount, such as marketing, fiction writing, and other areas where novelty and diversity are valued, base models may be more suitable. On the other hand, aligned models may be preferred when safety and consistency are the primary concerns, such as in customer support or content moderation.

The remainder of this paper is structured as follows: Section 2 provides background information on LLMs, RLHF, and their applications in marketing. Section 3 presents our experiments comparing the behavior of base and aligned models, followed by a discussion of the results and their implications in Section 4. Finally, Section 5 concludes the paper and outlines future research directions.

## 2.    Literature Review

RLHF has emerged as a promising technique to align LLMs with human preferences and values. However, recent research has highlighted several limitations and potential unintended consequences of RLHF, including scalability and efficiency concerns due to its reliance on human annotators (Lee et al., 2023; Yuan et al., 2023), variability and potential bias in human feedback affecting the quality and consistency of the process (Yu et al., 2023), vulnerability to manipulation by adversarial annotators leading to security breaches and ethical concerns (Wang et al., 2023), and alignment challenges such as objective mismatch and length bias (Lambert and Calandra, 2023; Shen et al., 2023).

Despite these limitations, LLMs have shown significant potential in transforming various aspects of marketing and business. They can automate and accelerate time-consuming tasks such as text generation, summarization, and content creation, leading to increased productivity and efficiency in marketing and business operations (Head et al., 2023). LLMs also enhance customer interaction by providing personalized and context-aware responses, which can improve customer satisfaction and engagement, particularly in customer service and support functions (Franceschelli and Musolesi, 2023). Furthermore, by analyzing large volumes of data, LLMs can generate valuable market insights, helping businesses understand customer preferences, market trends, and competitive landscapes, which can inform strategic decisions and marketing campaigns (Eloundou et al., 2023).

However, while the literature has explored the applications and limitations of LLMs in marketing and business contexts, there is a notable gap in understanding how the RLHF process affects the creativity and variation in the models' outputs. This is a crucial aspect for marketers and professionals who rely on LLMs for creative tasks. Understanding the trade-off between alignment with human preferences and the preservation of creative diversity in the generated content can have significant implications for the effectiveness and engagement of marketing initiatives.

## 3.    Experiments

Remember that our goal is to evaluate and contrast the diversity of texts produced by base models and their aligned counterparts. We focus on both the short-term (syntactic) and long-term (semantic) variations in model outputs using the Llama-2 language models. Meta has made both the base models[1] and their corresponding aligned versions[2] publicly available, making them an ideal choice for this study. Therefore, comparisons are made between Llama-2-7B-text (the "base" model) and Llama-2-7B-chat (the "aligned" model) where 7B refers to the parameter size of the LLM. These models are currently highly favored within the open-source community[1]. Their widespread use is partly attributed to the affordability of finetuning them.

---

[1] Referred to by Meta as "text" models.
[2] Referred to by Meta as "chat" models.

We conduct three experiments to examine the effects of the alignment process on model creativity and diversity. Experiment 1 serves as a concrete example of the differences in creativity between the base and aligned models, while Experiments 2 and 3 investigate the underlying mechanisms that contribute to these differences.

Each LLM is given an initial prompt which must be completed for a maximum[2] of $n_{\text{predict}}$ tokens[3]. LLMs typically generate one token at a time. At each step, the LLM produces a set of logits over the potential next tokens. These logits are then normalized to sum up to 1 using the softmax function, and one token is sampled randomly according to its probability:

$$\Pr(tok_i) = \frac{\exp(\text{logit}(tok_i)/T)}{\sum_i \exp(\text{logit}(tok_i)/T)} \qquad 3.1$$

where $tok_i$ is token number $i$ in the LLM's vocabulary[4] of tokens and $T \in (0,1]$ is a parameter called *temperature* which controls the "softness" of the probability distribution[5]. In our experiments we choose $T = 1.0$ for maximum response variation.

## 3.1. Experiment 1: Customer Persona and Review Generation

In this experiment, we generate customer personas using both the base and aligned models. For each model, we create 100 unique customer personas with the following attributes: first name, last name, gender, age, nationality, ethnicity, and personality type, according to the Myers-Briggs test ("Myers–Briggs Type Indicator," 2024). Additionally, each simulated customer writes a review for a hypothetical product: "*A coffee machine that connects to your smartwatch so it keeps your coffee warm if you are far away from it.*"

---

[1] Currently, there are more than 15,200 variants of Llama-2 models on the HuggingFace website: See https://huggingface.co/models?sort=trending&search=Llama-2

[2] LLMs may stop generating tokens even before reaching the $n_{\text{predict}}$ limit if they encounter an end-of-sequence (EOS) symbol. This symbol is `</s>` for Llama-2 and `<|im_end|>` for ChatML models (such as OpenAI's GPT-4).

[3] That is, the LLM continues our given prompt much like the autocomplete feature on smartphones. In Experiments 2 and 3, no further preprocessing is required. But in Experiment 1, we should format the prompt according to the chat template used during the Supervised Fine-Tuning (SFT) of the model ("Llama 2 Prompt Template," 2023).

[4] For Llama-2, the vocabulary size is 32,000 tokens (Touvron et al., 2023).

[5] High values of $T$ lead to more uniform and softer distributions, meaning that the LLM is more likely to generate creative and diverse outputs. Low values of $T$, on the other hand, result in peaked distributions with most of the probability mass concentrated on one or few tokens. This could lead to high confidence (but less variation) in the model's outputs.

To analyze the results, we first generate word clouds of the first and last names to visualize the diversity and variation in the names generated by each model. We then plot the distributions of ages, genders, and review lengths to compare the variety of these attributes between the base and aligned models.

Next, we compute and plot the distribution of sentiment polarity for the reviews using VADER[1], a sentiment analysis algorithm that assigns scores between $-1$ (negative) and $+1$ (positive) to texts (Hutto and Gilbert, 2015). This step allows us to compare the range of sentiments expressed in the reviews generated by each model.

To understand the variety in review semantics generated by the base and aligned models, we extract the *embeddings* of each sentence in the customer reviews. Embeddings are dense vector representations of words or texts that capture their semantic meaning in a high-dimensional space. The intuition behind using embeddings is that they can capture the semantic similarity of the generated outputs, even if the outputs differ at the token-level[2]. To calculate the embeddings, we use Sentence-BERT (SBERT), a state-of-the-art framework for text and image embeddings[3] (Reimers and Gurevych, 2019). Given an input text, SBERT converts it to a 384-dimensional vector in the embedding space.

We determine the optimal number of k-means ("*k*-means clustering," 2024) clusters for the embeddings of each model and visualize the clusters using t-distributed Stochastic Neighbor Embedding (t-SNE) (van der Maaten and Hinton, 2008), a technique for dimensionality reduction that is particularly well-suited for the visualization of high-dimensional datasets. By applying t-SNE to the 384-dimensional embeddings, we can project them onto a 2D space while preserving the local structure of the data, making it easier to identify clusters and patterns. If the LLM's outputs form tight clusters in the t-SNE visualization, it would indicate a lack of semantic diversity and creativity. Vice versa, if the outputs are spread out, it suggests higher variation.

---

[1] Valence Aware Dictionary and sEntiment Reasoner

[2] For example, it could be that the LLM generates texts that are syntactically different because it has many alternative wording choices. However, these choices may nevertheless all lead to semantically similar outputs. In such cases, the generated texts will have embeddings that are very close in the embedding space, indicating a lack of variety in semantics and, consequently, a lack of creativity in model output. On the other hand, if the model generates outputs with more diverse embeddings, it would suggest a higher level of semantic variety and, therefore, more creativity in the long run.

[3] SBERT uses a pre-trained BERT model (Devlin et al., 2019) to encode the input text into a fixed-size vector representation. The BERT model is an encoder-only transformer trained on a large corpus of text data using a self-supervised learning approach, which allows it to learn rich, contextual representations of words and sentences. SBERT fine-tunes the pre-trained BERT model on a sentence similarity task, resulting in a model that can generate high-quality sentence embeddings.

## 3.2. Experiment 2: Semantic-level Variation in LLM Outputs

In this experiment, we examine the long-term semantic diversity of the base and aligned models when given a simpler task that does not explicitly require creativity. We set the initial prompt to "*Grace Hopper was*" with $n_{\text{predict}} = 128$, allowing the model to generate long sequences of tokens. The goal is to assess the model's ability to recite a historical fact about Grace Hopper in various ways, focusing on its capacity in expressing the same information using different wordings and sentence structures.

We generate 200 outputs from each model and calculate their embeddings using SBERT as in Experiment 1 and reduce the dimensionality using t-SNE. However, unlike Experiment 1, we now calculate the embeddings for the entire generations rather than for individual sentences, capturing the holistic semantics of the model output.

To complement the t-SNE visualization, we calculate the cosine similarity scores between pairs of embedding points using the TF-IDF (Term Frequency-Inverse Document Frequency) vectorizer (Salton and Buckley, 1988; Sparck Jones, 1972). TF-IDF is a widely used technique in natural language processing that converts the generated texts into numerical vectors where each dimension represents a unique word in the corpus, and the value of each dimension is the TF-IDF weight of the corresponding word[1]. These vectors can then be used to calculate the cosine similarity between pairs of generated outputs, providing a quantitative measure of their semantic similarity.

## 3.3. Experiment 3: Syntactic Diversity in LLM Outputs

In this experiment, we investigate the short-term token-level probabilities of the base and aligned models. We hypothesize that the difference in semantic diversity between the two models could be due to the aligned model's inability to assign more spread-out probabilities to tokens, resulting in certain token trajectories being blocked or unavailable. In other words, syntactic diversity is a necessary condition for semantic diversity.

To test this hypothesis, we set the initial prompt to "*Steve is the CEO of a startup company*" with $n_{\text{predict}} = 64$ so the model generates a background story for Steve. For each generated token, we extract the top five predicted tokens according to their probability (see Eq. 3.1) and calculate their Shannon entropy[1] (Shannon, 1948) as follows:

$$H_n = -\sum_{j=1}^{5} \Pr(tok^j) \log_2(\Pr(tok^j)) \qquad 3.2$$

where $H_n$ is the entropy of the $n$-th generated token ($1 \leq n \leq n_{\text{predict}}$) and $tok^j, j \in [1,5]$ are the top five tokens predicted by the LLM[2]. The intuition is that a more creative model will generate a wider variety of tokens, resulting in a higher average entropy across its predictions. Conversely, a less creative model is expected to have a more skewed probability distribution, favoring certain tokens over others, leading to lower entropy values.

---

[1] Calculated based on its frequency within the document and its rarity across the entire corpus.

For each completion, we compute the average entropy of the generated tokens. The mean and standard deviation of these average entropies is then calculated across 100 completions. This approach allows us to compare the average token variation between the two LLMs while reducing the impact of outliers or inconsistencies in individual completions.

## 4. Results

### 4.1. Customer Persona and Review Generation

We begin by analyzing the word clouds of the first and last names generated by the base and aligned models (Figure 1). While the aligned model heavily favors few names such as "Emily" and "Samantha" for first names and "Jones" and "Wang" for last names, the base model produces a much wider variety of names. This observation suggests a potential lack of creativity in the aligned model resulting from the RLHF process.

---

[1] The Shannon entropy is a measure of the uncertainty or information content of a random variable, and can be thought of as the level of randomness in the model's predictions at each step. A higher Shannon entropy indicates more uncertainty or a more uniform probability distribution over the predicted tokens, while a lower entropy suggests the model is more confident or has a more skewed probability distribution favoring certain tokens.

[2] The Shannon entropy in Eq. 3.2 is measured in *bits*, with a lower-bound of 0 bits corresponding to complete certainty (i.e., the model assigns a probability of 1 to a single token and 0 to all others) and an upper-bound of $\log_2(5) \approx 2.32$ bits, which occurs when the model assigns equal probabilities to all 5 tokens under consideration.

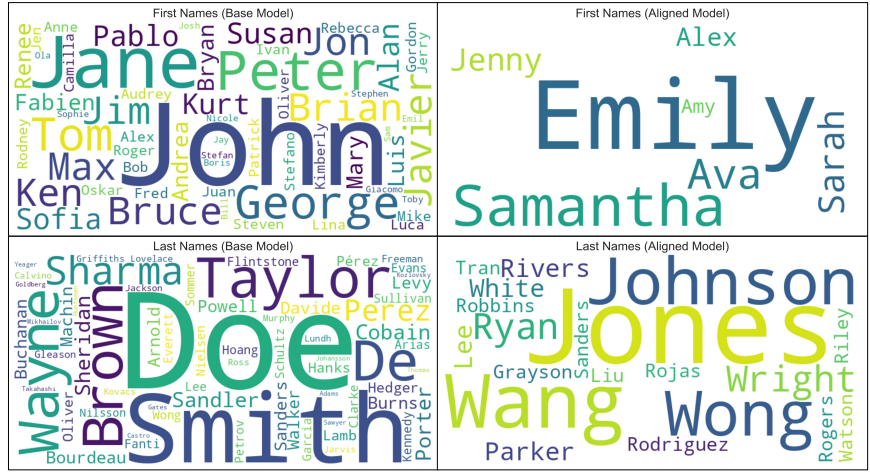

Figure 1. Word cloud of first and last names of the synthetic customers generated by the base and aligned models.

Next, we examine the diversity in the demographic attributes of the simulated customers, including nationality, ethnicity, personality type, and age. Figure 2 shows the distribution of nationalities generated by the base and aligned models. The base model generates a wide range of nationalities, with American, British, and German being the top three. In contrast, the aligned model only generates three nationalities: American (highest percentage), Chinese, and a small percentage of Mexican.

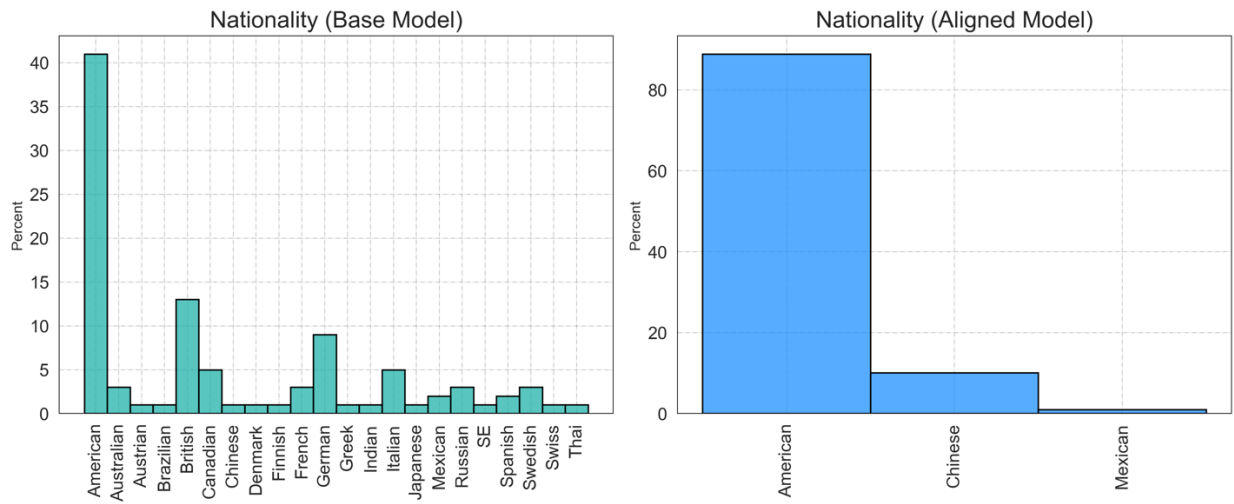

Figure 2. The distribution of nationalities of the customers generated by the base and aligned models.

A similar trend is observed in the distribution of ethnicities (Figure 3). The base model generates various ethnicities, including White, Asian, Black, Latino, and even some minorities such as Ashkenazi Jewish. On the other hand, the aligned model primarily generates White and Asian, with a smaller percentage of Latino.

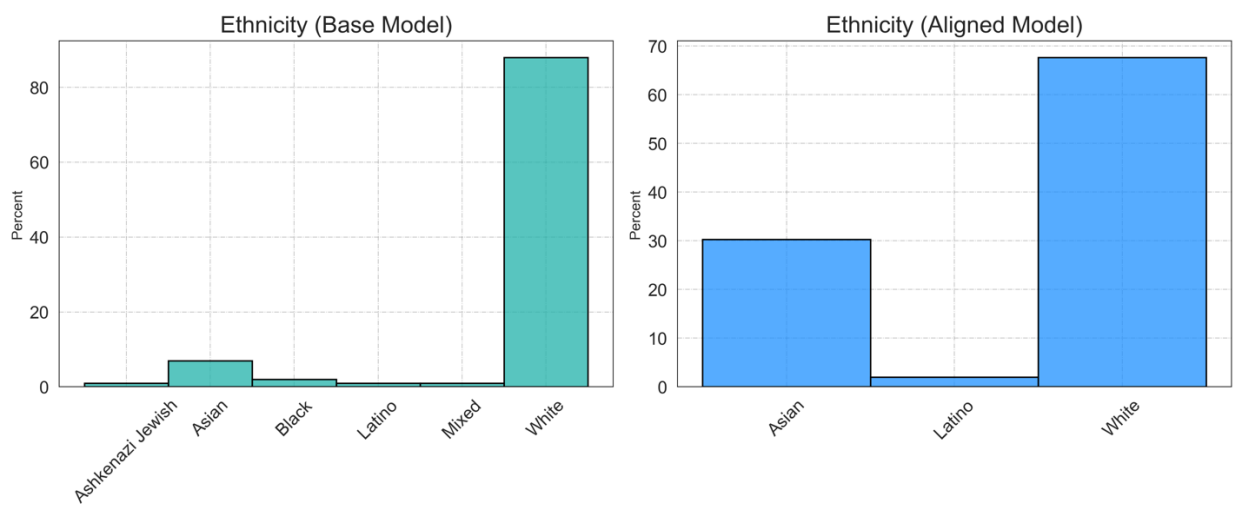

Figure 3. The distribution of ethnicities of the customers generated by the base and aligned models.

When analyzing the distribution of personality types (Figure 4), we find that the base model generates all 16 personality types defined by the Myers-Briggs test. In contrast, the aligned model only generates six personality types, indicating a significant reduction in diversity.

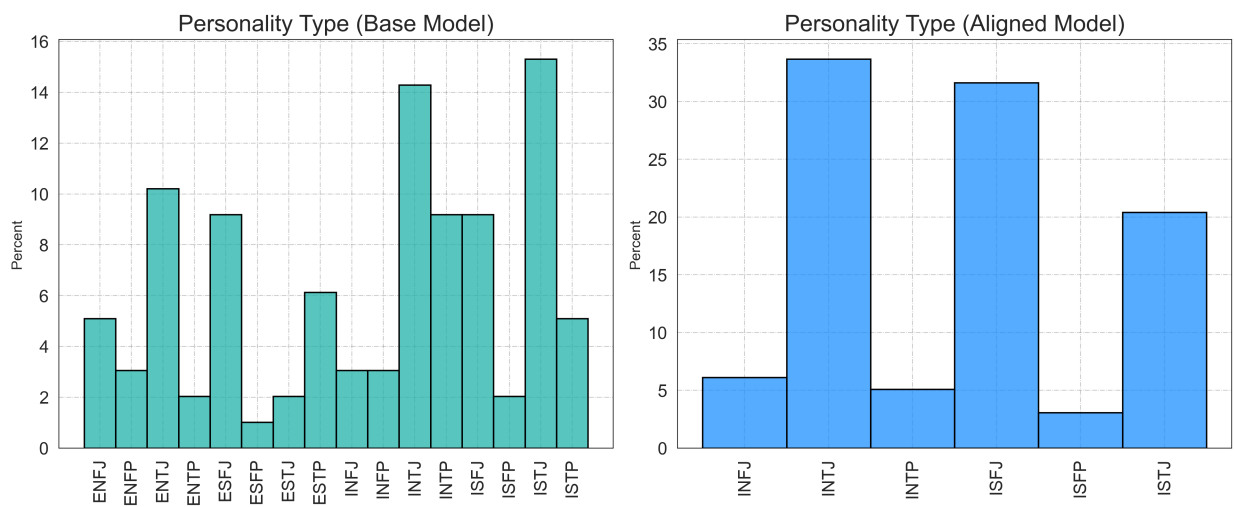

Figure 4. The distribution of personality types of the customers generated by the base and aligned models.

The distribution of ages for the simulated customers (Figure 5) further highlights the differences between the two models. The base model's age distribution resembles a normal distribution, spanning from below 10 years old to nearly 70 years old, with the majority centered around 30. The aligned model, however, selects ages within a narrow range, with a strong preference for age 32 and a few other ages between 28 and 35. Notably, the aligned model does not select any ages above 35 or below 28, indicating a limited capability in generating diverse age values.

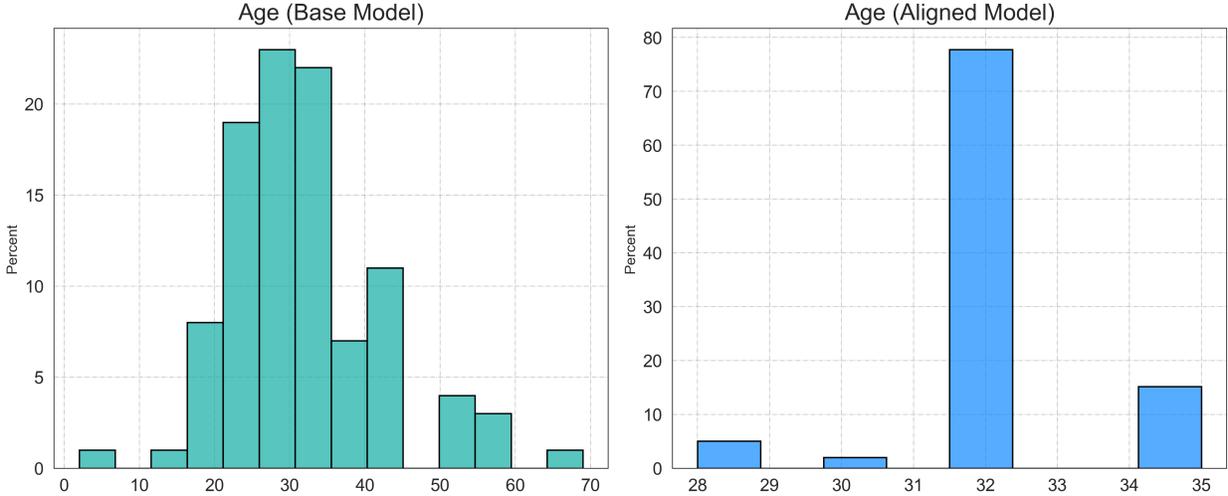

Figure 5. The distribution of ages of the customers generated by the base and aligned models.

Finally, the distribution of customer gender (Figure 6) shows that the base model generates approximately 80% male and 20% female customers, while the aligned model generates nearly 100% female customers, with a negligible number of males.

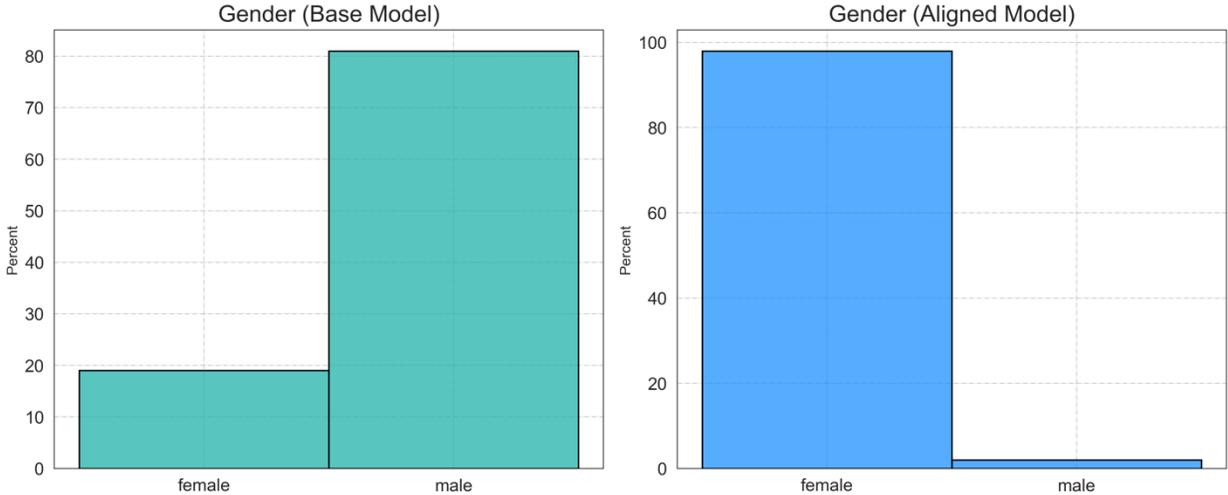

Figure 6. The distribution of genders of the customers generated by the base and aligned models.

**The Effect of Token Noise:** It is important to note that the goal of this paper is not to study bias or demographic diversity in language models, as there is already extensive research on this topic. As discussed in (Mohammadi, 2024), the effect of *token noise* can significantly influence the distributions of preferences generated by language models, depending on the specific prompt used. Language models do not have a single set of preferences. Rather, they are data generating processes that generate distributions of distributions based on the input prompt. Therefore, the focus of this study is not on the change in the distribution of specific demographic attributes, but rather on the overall variety and diversity of the generated outputs.

Moving on to the analysis of the product reviews, we first examine the distribution of review lengths (Figure 7). The aligned model generates significantly longer reviews compared to the base model, with an average length of 457 characters for the aligned model and 109 for the base model. However, the length of the review alone does not provide much insight into the content and diversity of the reviews. To gain a deeper understanding of the review sentiments, we analyze the polarity of the reviews (Figure 8). The base model covers a wider range of sentiments, from $-0.75$ to $+0.97$, with most of the distribution skewed towards positive sentiments. In contrast, the aligned model concentrates almost entirely around $+1$, indicating that the sentiments of the reviews generated by the aligned model customers are overwhelmingly positive about the product. While positive reviews are desirable, generating a diverse set of customer experiences, including some negative or neutral reviews, is important for realistic market simulations.

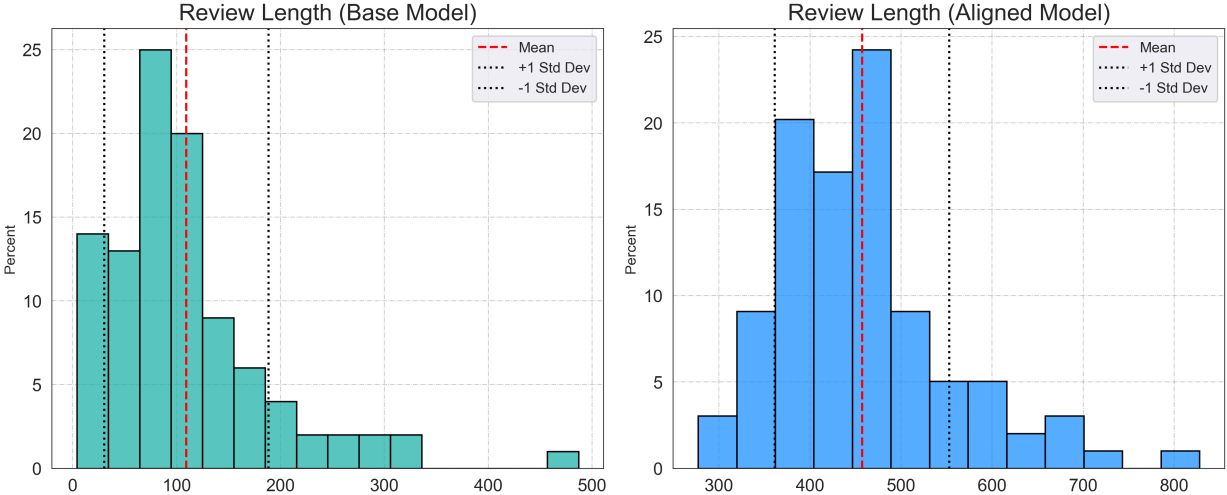

Figure 7. The distributions of the length of the reviews by simulated customers of the base and aligned models.

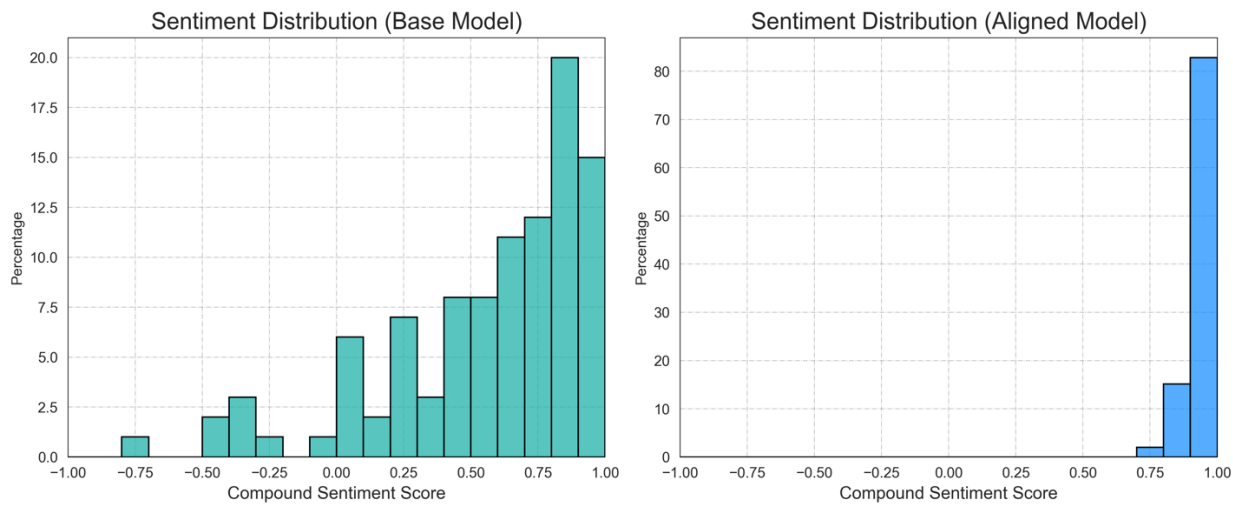

Figure 8. The distributions of the sentiments of the reviews generated by the simulated customers produced by the base and aligned models.

To quantify the diversity in the review content, we calculate the embeddings of each sentence in the reviews and cluster them using k-means clustering. The optimal number of clusters is determined to be 14 for the base model and 6 for the aligned model, with a perplexity of 30 chosen for both t-SNE plots (Figure 9). The t-SNE visualization reveals that the sentences from the aligned model reviews form distinct clusters, whereas the base model sentences exhibit more spread and heterogeneity. This finding suggests that the aligned model generates reviews with similar sentence structures, word choices, and overall content, while the base model produces more diverse and varied reviews.

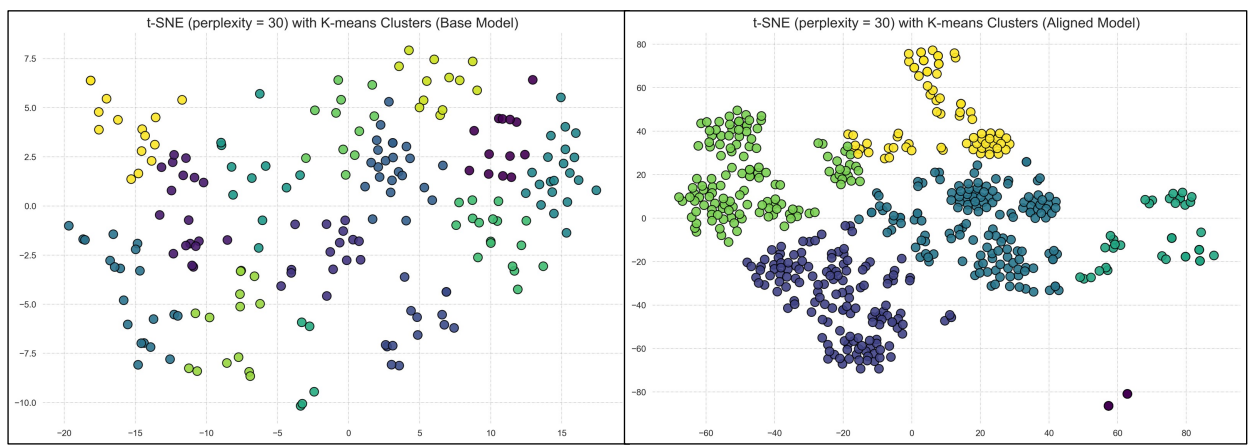

Figure 9. The t-SNE plot of the embeddings of the product reviews generated by simulated customers of the base and aligned models. Colors indicate clusters.

Table 2 and Table 3 present sample sentences from each cluster for both the base and aligned models. The aligned model clusters exhibit repetitive patterns, such as sentences focused on emojis, phrases like "highly recommend" or "I highly recommend it", and verbatim repetitions of sentences such as "this machine is a game changer". Furthermore, this model generates sentences with similar structures like "as a busy professional […]" or "as someone who is always on the go […]." In contrast, the base model clusters do not display such repetitive patterns, indicating a higher level of semantic and syntactic diversity in the generated reviews.

The results of Experiment 1 highlight the significant differences in the variety of demographics and review content between the base and aligned models when generating simulated customers for a practical marketing application. The aligned model, which has undergone the RLHF process to reduce bias and toxicity, appears to have lost its ability to generate diverse outputs. This finding motivates the need to investigate the underlying causes of this creativity loss, which will be explored in the subsequent experiments.

## 4.2. Semantic Diversity and LLM Output Embeddings

To illustrate the differences in the semantic-level variation of the base and aligned models, we first present a sample of the generated outputs for the initial prompt "*Grace Hopper was*" (Table 4). The sample outputs demonstrate that while both models generate factually correct information about Grace Hopper, the base model exhibits more diversity in its wording and sentence structures compared to the aligned model.

The t-SNE visualization of the embedding space (Figure 10) reveals distinct clustering patterns for the base and aligned models. The embedding points of the base model are scattered and spread out, indicating a higher level of semantic diversity in the generated outputs. In contrast, the aligned model's embeddings form four distinct clusters (See Table 5 for an example of each), with empty spaces between them, suggesting that the aligned model tends to stick to certain embeddings or generations and expresses the information in a limited number of ways.

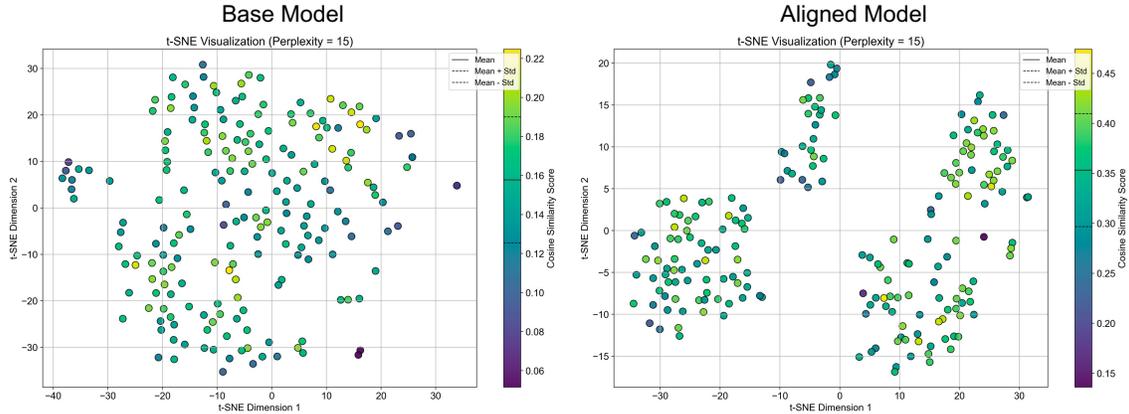

Figure 10. The t-SNE visualization of the embedding space of LLM outputs and their cosine similarity scores.

The cosine similarity analysis using TF-IDF vectorization provides further evidence of the differences in semantic diversity between the two models. The average cosine similarity score for the base model is 0.16 (STD = 0.03), while the aligned model has a higher average similarity score of 0.35 (STD = 0.06). These scores are visualized on the plots using a color scale.

The results from Experiment 2 demonstrate that the base model exhibits higher variation compared to the aligned model at the semantic level, as evidenced by the more diverse embeddings of its generated outputs. This finding confirms the results of Experiment 1, further supporting the hypothesis that the alignment process constrains the creative capabilities of language models.

### 4.3. Syntactic Diversity and Average LLM Entropy

The results obtained from Experiment 3 reveal a significant difference in the average entropy between the base model and the aligned model. The base model exhibits a higher mean entropy of 1.48 (STD = 0.17), while the aligned model has a lower mean entropy of 0.96 (STD = 0.12). This difference is illustrated in the box plot in Figure 11. To emphasize the contrast between the two models, we introduce the terminology "hot model" for the base model and "cold model" for the aligned model. This nomenclature reflects the inherent differences in their token-level entropies and creativity.

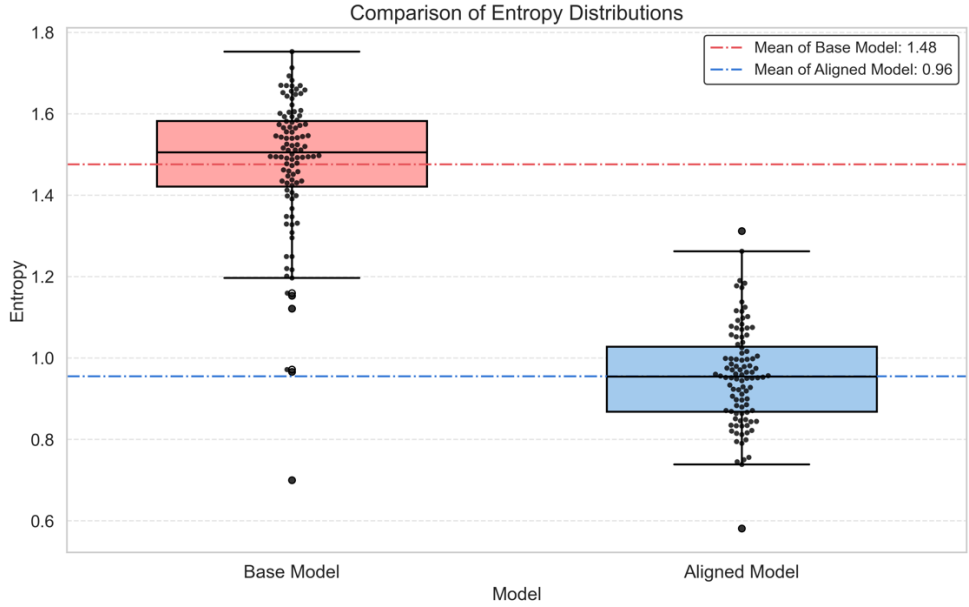

Figure 11. The boxplot of output entropies for the base and aligned LLMs

To further illustrate the disparity between the two models, Figure 12 provides a side-by-side comparison of a sample generation from each model (where $n_{\text{predict}} = 16$), along with their corresponding entropy bar plots. The base model consistently displays higher entropy values compared to the aligned model, reinforcing the notion that the base model exhibits greater token-level variation.

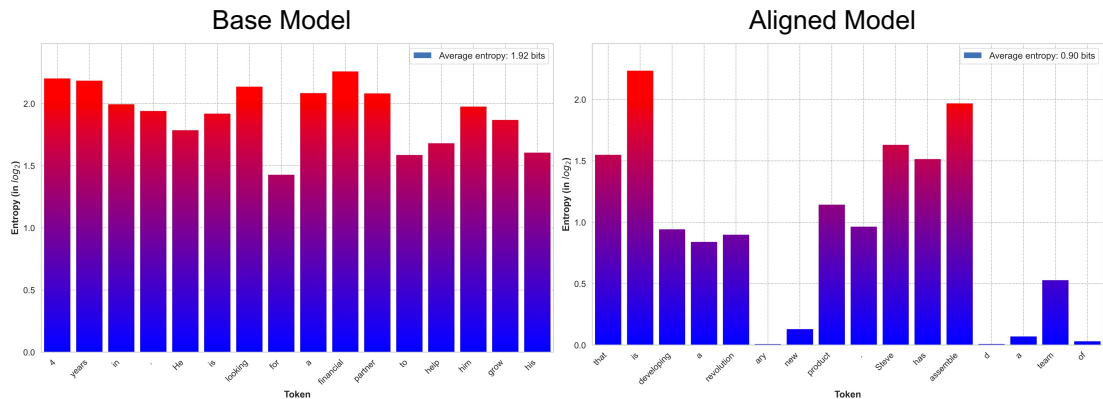

Figure 12. A Sample of Token Entropies.

In addition to the entropy analysis, Figure 13 presents stacked bar plots of the probability distributions for each predicted token in the sample generation. The base model's probability distributions appear more spread out, indicating that it assigns more evenly distributed probabilities to different tokens. Consequently, when the base model randomly samples from these distributions, it has a higher likelihood of generating diverse token trajectories. In contrast, the aligned model's probability distributions are more peaked, with most of the probability mass concentrated on one or two tokens. As a result, when the aligned model samples from these distributions, it tends to generate the same tokens more frequently, leading to less diverse output.

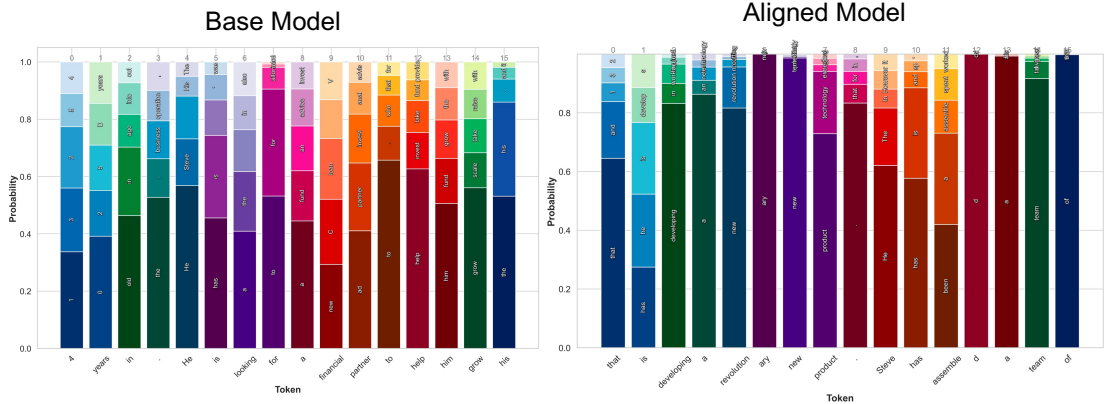

Figure 13. A Sample of Token Probabilities.

These findings highlight the significant difference between the base model and the aligned model in terms of token-level variations, reinforcing our hypothesis that syntactic diversity is a necessary condition for semantic diversity. Although high token-level variation does not guarantee semantic diversity, it is a necessary condition, and the aligned model, with its low token-level entropy, is incapable of producing semantically diverse outputs. This suggests that the RLHF process converts the LLM into a more deterministic algorithm that lacks the capacity to explore diverse sets of token trajectories.

## 5. Discussion

### 5.1. Attractor States and Model Creativity

Experiment 2 revealed that the aligned model's outputs form four distinct clusters, suggesting that it can only describe Grace Hopper in four different ways. By analyzing the generated paragraphs, we observe that the model follows certain patterns in terms of word choice, sentence structure, and the overall content being discussed.

An intriguing question is: What happens if we intentionally **perturb** the trajectory of tokens or the path that the model takes when generating outputs? To explore this, we take one of the four distinct ways the aligned model described Grace Hopper and modify the first sentence by changing the last word from "*was*" to "*was not*" (Figure 14). We then append this to the initial prompt "*Grace Hopper was*" to obtain a new initial prompt such as "*Grace Hopper was born on November 9, 1906 in New York City. She was not*".

Interestingly, when presented with this perturbed prompt, the aligned model gracefully[1] finds its way back to one of its own completion distributions.

Notice that the aligned model's outputs in Figure 14 are predominantly green, reflecting its tendency to generate high-probability tokens as discussed in Experiment 3. Nonetheless, when we perturb the initial prompt of the aligned model, we can witness its struggle to find the appropriate tokens to justify the perturbation. The color of the first few tokens generated by the model may be red, orange, or yellow, indicating lower probabilities and a challenge in steering the completion back to its original path. However, once the model successfully navigates back to this "familiar" state, the token colors return to green, signifying a return to high-probability completions.

This phenomenon is reminiscent of *attractor states* in system dynamics (Janus, 2022). Attractor states are regions in a system's phase space towards which the system tends to evolve, even when slightly perturbed. In the case of the aligned model, the four distinct ways of describing Grace Hopper can be seen as attractor states[2]. This makes the aligned model resemble more a goal-directed agent—which seems to have already decided what it is going to say—than an autoregressive model capable of generating various completions for its initial input. While the behavior of the aligned model ensures consistency and coherence in the model's outputs, it also highlights a potential limitation in terms of creativity. For truly creative models, we desire the ability to explore diverse ways of expressing ideas and generating novel concepts, rather than being confined to a limited set of attractor states.

It is important to note that the attractor states observed in the aligned model are different from the behavior of models at low temperature. We observe these attractors even at high temperatures (e.g., $T = 1$) for the aligned model, whereas we do *not* observe similar attractors at low temperatures for the base model. This suggests a qualitative difference between the aligned model and the base model which does not get trapped in attractor states and thus can generate a wider variety of outputs. In the next section, we will delve deeper into the reasons behind this phenomenon, drawing insights from the description of the RLHF process by Meta in (Touvron et al., 2023).

---

[1] Pun intended.

[2] When the model is nudged away from these states through perturbations in the prompt, it finds its way back to the attractor, much like how masses are drawn towards black holes in space.

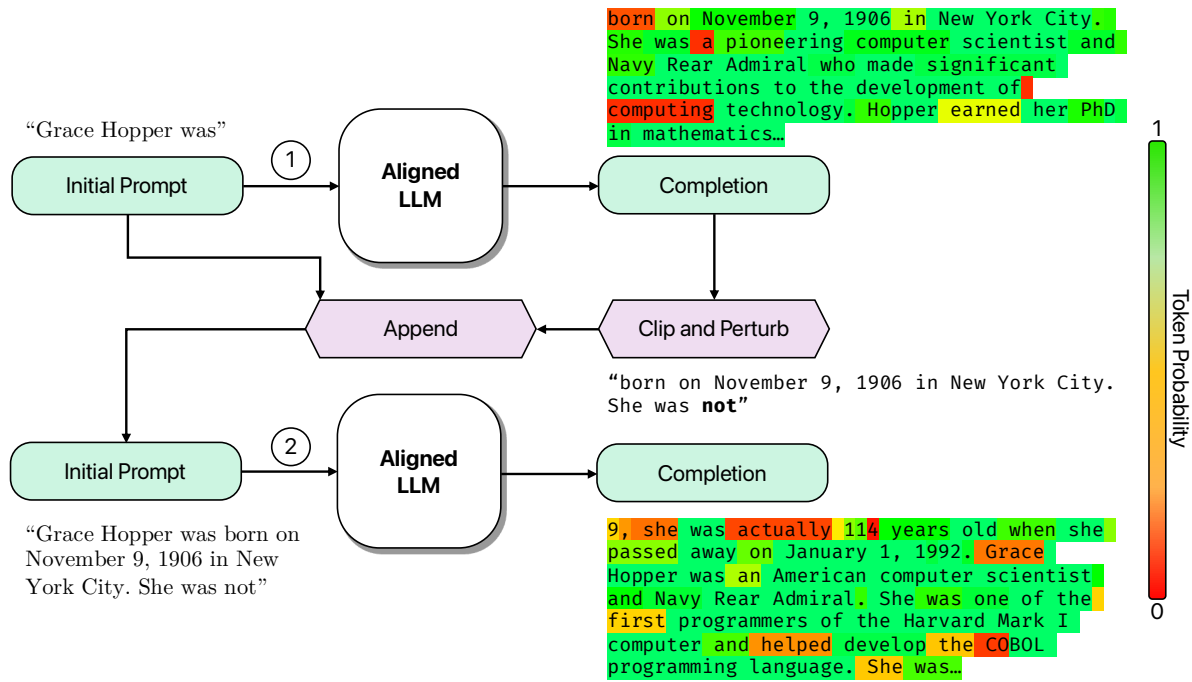

Figure 14. Slightly nudging the aligned LLM out of its completion distributions. The aligned LLM finds a way to get back to one of its own completion distributions.

## 5.2. Why Alignment Reduces Creativity: Insights from RLHF and PPO

The attractor states observed in the aligned Llama-2 models (Section 5.1) can be attributed to *mode collapse[1]*, a common problem in reinforcement learning (RL) where the agent gets stuck in a limited set of behaviors or outputs. Mode collapse occurs when the agent's policy converges to a suboptimal solution that maximizes the immediate reward but fails to explore potentially better strategies. This issue also plagues Reinforcement Learning from Human Feedback (RLHF), a popular technique for aligning language models with human preferences.

In RLHF, human preference data is collected by asking human annotators to compare and rank model-generated responses. This data is then used to train a reward model that estimates the quality of a response based on its alignment with human preferences. Finally, the pretrained language model is fine-tuned using the reward model through an RL algorithm, such as Proximal Policy Optimization (PPO)[1].

---

[1] Mode collapse occurs when a generative model, like a Generative Adversarial Network (GAN), produces a limited set of outputs repeatedly. This happens because the generator over-optimizes for a specific discriminator, which fails to adapt, leading to the generator producing the same outputs. This cyclical failure results in a lack of variety in the generated outputs ("Common Problems," 2024).

In PPO, the policy (i.e., the language model) is updated based on the *advantage function* $A^\pi(s, a)$ defined as follows:

$$A^\pi(s, a) := \mathbb{E}_{s' \sim T(s,a)}[R(s, a, s') + \gamma v^\pi(s')] - v^\pi(s) \qquad 5.1$$

where $s$ represents the current state, $a$ represents the action taken in state $s$, $s'$ is the next state after taking action $a$ in state $s$, $T(s, a)$ is the transition function that determines the probability of reaching state $s'$ from state $s$ by taking action $a$, $R(s, a, s')$ is the reward received when transitioning from state $s$ to state $s'$ by taking action $a$, $\gamma$ is the discount factor, and $v^\pi(s)$ is the state-value function, which estimates the expected return starting from state $s$ and following policy $\pi$.

Intuitively, $A^\pi(s, a)$ measures how much better a specific action $a$ (i.e., generating a particular response) is compared to the average action taken by the current policy $\pi$ given the current state $s$ (i.e., the user's prompt). To illustrate how this could lead to arbitrarily high logits on certain LLM responses, let us consider a simple example where a model is asked to generate a name for a new product. The prompt is "*Create a name for a new chat bot powered by generative AI*". During the RLHF process, the model generates two responses: "*Jeepiti*" and "*Chats and Giggles*", with rewards of 1.0 and 0.4, respectively[2]. Assume that at the beginning, the model generates these responses with equal probability, and the value function $v^\pi(s)$ is set to zero. The policy update rule is: Add $A^\pi(s, a)$ to the logits (log-odds) of action $a$. For instance, if $\pi_0(s, \text{Jeepiti}) = 0.5$ and $A^{\pi_0}(s, \text{Jeepiti}) = 1.0$, then the probability of "*Jeepiti*" increases from 0.5 to 0.73[3]. The following table demonstrates how the advantage function oscillates under policy updates, extracting an unbounded amount of reinforcement from a single action (in this case, "*Jeepiti*")[4]:

---

[1] For a more detailed explanation of the RLHF process, please refer to the Appendix A2.

[2] The reward given by the reward model during the RLHF process. The reward model estimates human preferences.

[3] Suppose in the beginning $\pi_0(s, \text{Jeepiti}) = \pi_0(s, \text{Chats and Giggles}) = p = 0.5$. The logits (log-odds) of both actions are: $\log(p/(1-p)) = \log(1) = 0$. Adding the advantage $A^\pi(s, \text{Jeepiti}) = 1$ to the logit of "*Jeepiti*", we get the new logit: logit(Jeepiti) $= 1 + 0 = 1$. Converting the logits back to probabilities using the softmax function gives: $\pi_1(s, \text{Jeepiti}) = e^1/(e^0 + e^1) \simeq 0.73$ and $\pi_1(s, \text{Chats and Giggles}) = 1 - 0.73 = 0.27$.

[4] This is a simplified example. The actual PPO algorithm works differently and incorporates additional mechanisms such as clipping and early stopping based on KL divergence thresholds to mitigate mode collapse.

Table 1. An illustrative example of mode collapse during RLHF

| $t$ | LLM Response $a$ | Reward $R(a\|s)$ | Advantage $A^\pi(s,a)$ | $v_t^\pi(s)$ | $\pi_t(s, \text{Jeepiti})$ | $\pi_t(s, \text{Chats and Giggles})$ |
|---|---|---|---|---|---|---|
| 0 | — | — | — | 0.0 | 0.5 | 0.5 |
| 1 | *"Jeepiti"* | 1.0 | $(1+0) - 0 = 1$ | 1.0 | 0.73 | 0.27 |
| 2 | *"Chats and Giggles"* | 0.8 | $(0.4+0) - 1 = -0.6$ | 0.4 | 0.83 | 0.17 |
| 3 | *"Chats and Giggles"* | 0.8 | $(0.4+0) - 0.4 = 0$ | 0.4 | 0.83 | 0.17 |
| 4 | *"Jeepiti"* | 1.0 | $(1+0) - 0.4 = 0.6$ | 1.0 | **0.93** | **0.07** |

As we can see, the model is receiving substantial positive reinforcement for the response *"Jeepiti"*, causing its internal circuits (i.e., neural network weights) to be reshaped in a way that favors this response. This behavior is undesirable because ideally we want $\pi_t(s, a)$ to be proportional to the reward of $a$, meaning that the reward should update the policy only by a finite amount. Moreover, the model can become trapped in a local optimum. For instance, if the goal is to have the model respond with other names for the chatbot and a reward of 1.5 is provided for saying *"Chad-Chat"*, exploration issues might prevent the model from ever producing this response during training. As noted by (TurnTrout and MichaelEinhorn, 2023), this issue is exacerbated by the fact that PPO actively updates the policy against actions that do not outperform the current (on-policy) value function $v_t^\pi(s)$. This process tends to discourage exploration, as it penalizes actions that do not directly increase the estimated value.

Both the GPT-3 (Ouyang et al., 2022) and Llama-2 (Touvron et al., 2023) models used PPO for RLHF, as well as a KL penalty[1] in the reward function to encourage the updated policy to stay close to the original policy:

$$R(a|s) = \widetilde{R}_c(a|s) - \beta D_{\text{KL}}\big(\pi(\cdot \,|s) \,\|\, \pi_0(a|s)\big) \qquad 5.2$$

where $\widetilde{R}_c$ is the combined reward from the helpfulness and safety reward models, $\beta$ is the KL penalty coefficient, $\pi$ is the updated policy, and $\pi_0$ is the original policy. For Llama-

---

[1] The Kullback-Leibler (KL) divergence, denoted by $D_{\text{KL}}(p\|q)$, is a measure of how probability distribution $p$ is different from (a reference) probability distribution $q$. For discrete distributions $p, q$, we have:
$$D_{\text{KL}}(p\|q) := \sum_x p(x)[\ln p(x) - \ln q(x)]$$

2, relatively small values for *β* were used[1] to balance the trade-off between allowing the model to optimize for the reward and maintaining stability during training.

Nonetheless, this definition of the reward function implicitly assumes that human preferences can be encapsulated solely as a function of the prompt (*s*) and the generated response (*a*). However, human preferences can be contingent on the *distribution* of model outputs (particularly in scenarios where human evaluators are forced to select one of two model responses during RLHF, as in the case with Llama-2). This constraint poses a significant challenge for the model to converge to a reasonable stochastic strategy, especially with only 4,000 labels used to RLHF the Llama-2 model (Touvron et al., 2023).

Indeed, despite the use of PPO with clipping and KL divergence penalty, both GPT-3 and Llama-2 models still exhibit mode collapse issues. (Ouyang et al., 2022) found that even increasing the KL penalty coefficient *β* by a factor of 100 was not sufficient to recover performance on public NLP datasets, and it caused a significant drop in the validation reward. As an alternative, they proposed mixing the pretraining gradients into the PPO gradients during RLHF fine-tuning (a technique they called "PPO-ptx"). While this approach performed better than increasing *β*, the authors noted that it still did not completely mitigate the performance regressions and could introduce undesirable behaviors if the pretraining data contained biases or toxicity. Furthermore, the persistence of mode collapse in GPT-3 models despite the use of PPO-ptx has been documented (Janus, 2022).

These findings suggest that the alignment process using RLHF might be fundamentally problematic and can lead to an "*alignment tax*"—a term used to describe the performance degradation observed in aligned models compared to their base counterparts (Lin et al., 2024). The limitations of the reward function and the difficulty in mitigating mode collapse through modifications to the PPO algorithm or hyperparameters highlight the need for alternative approaches.

As proposed by (Ouyang et al., 2022), a potentially more effective approach to reducing biases and toxicity in language models, while preserving their creative potential, could be to focus on the quality and diversity of the pretraining data itself. By carefully curating and filtering the pretraining data to ensure that the model learns from high-quality, diverse, and unbiased examples, we may be able to mitigate the need for extensive post-hoc alignment techniques like RLHF, which can inadvertently lead to mode collapse and reduced creativity.

---

[1] 0.01 for smaller models and 0.005 for larger models.

# 6. Conclusion

In this paper, we have investigated the impact of the Reinforcement Learning from Human Feedback (RLHF) alignment process on the creativity and output diversity of Large Language Models (LLMs). Our experiments, conducted using the Llama-2 series of models, have revealed that while RLHF is effective in reducing biases and toxicity in LLMs, it may inadvertently lead to a reduction in the models' creative potential, defined as the ability to generate outputs with high syntactic and semantic diversity.

We have taken a foundational approach to studying this problem by examining the issue at both the semantic and syntactic levels through three experiments. Experiment 1 demonstrated the impact of RLHF on creativity in a practical marketing context by comparing the diversity of customer personas and product reviews generated by base and aligned models. Experiment 2 investigated the semantic diversity of the models' outputs, revealing that aligned models form distinct clusters in the embedding space, indicating a fundamentally limited range of outputs compared to their base counterparts. Experiment 3 delved into the syntactic diversity, showing that aligned models exhibit lower entropy in token predictions. This suggest that the cause of this reduction in model creativity is the fact that many token trajectories become blocked during the RLHF process, i.e., the model loses its ability to produce certain tokens (their probability becomes almost zero), even if they have nothing to do with generating toxic or biased content. This makes aligned models function more like deterministic algorithms rather than creative generative models.

Furthermore, we have observed that the aligned model's outputs tend to gravitate towards specific "attractor states", a phenomenon related to *mode collapse* in reinforcement learning. This behavior highlights the challenges in preserving the creative potential of LLMs while aligning them with human preferences. In contrast, the base model exhibits greater flexibility and adaptability in its outputs.

The implications of these findings are significant for marketers and other professionals who rely on LLMs for creative tasks, such as copywriting, ad creation, and customer persona generation. The trade-off between consistency and creativity in aligned models should be carefully considered when selecting the appropriate model for a given application. In situations where creativity and diversity are paramount, base models may be more suitable, while aligned models may be preferred when safety and consistency are the primary concerns. Additionally, our results are important for those studying recommendation systems, as the insights can inform the development and optimization of these systems to balance creativity, diversity, and reliability effectively.

It is important to note that while base models offer greater creative potential, they are not directly usable in applications like chatbots. This is where techniques such as prompt engineering become increasingly important. Contrary to the belief that prompt engineering may become obsolete, our findings suggest that these techniques will be more crucial than ever in harnessing the power of base models for various applications. By carefully crafting input prompts that include instructions, examples, or constraints, users can guide the models' outputs and make them more suitable for specific use cases while still leveraging their creative potential.

A potential area for further investigation is the exploration of various parameters or configurations of the RLHF process, as higher computational costs and resource demands limited our ability to delve into these aspects. Future research could examine how different parameters influence the creativity and output diversity of aligned LLMs. Moreover, additional studies should analyze other unintended consequences of model alignment and RLHF to enhance our understanding of the trade-offs involved in practical applications of these models.

# Appendices

## A1.   Additional Tables

Table 2. Sample sentences from the clusters of review embeddings of the customers generated by the base model.

```
    Cluster 0:
    - Coffee is a great gift, this gift will keep you warm and let you share it with people
you love
    - It is really compact, and the temperature is perfect!
    - It is so great that I can keep my coffee warm, even if I am far away from it!
    Cluster 1:
    - I would recommend it to anyone.
    - This product is great.
    - I highly recommend this product.
    Cluster 2:
    - This machine is amazing!
    - I have an iPhone 6s and I could pair it with my phone.
    - This machine is the best one that I've ever bought.
    Cluster 3:
    - It has an app that works with your smartwatch and lets you control it with your watch
so you don't have to walk back to your coffee machine if you are away from it.
    - The smartwatch feature doesn't work either.
    - I like the fact that it can connect to my smartwatch
    Cluster 4:
    - This coffee machine is great.
    - This is the best coffee machine I have ever seen!
    - This coffee machine is great!
    Cluster 5:
    - I love it so much.
    - I saved so much time.
    - The design is also very nice!
    Cluster 6:
    - It doesn't keep my coffee warm if I'm far away from it.
    - It is easy to use, it keeps my coffee hot and it is really stylish.
    - It also saves money on the electricity bill as I never have to worry about whether the
coffee is still warm enough.
    Cluster 7:
    - It works great.
    - It looks sleek and it tastes great.
    - I have it with me everywhere I go.
    Cluster 8:
    - My only problem is that I have to recharge it twice a day because it has a battery
life of about 4 hours.
    - The only problem is that it is a bit noisy!
    - I can't control it.
    Cluster 9:
    - It's a nice idea to have your coffee stay warm, but the machine has a design issue:
it's too big to put on a table.
    - Amazing, it keeps my coffee warm no matter how far I am from the machine
    - The coffee is always cold when I get back to my office.
```

**Cluster 10:**
- It has everything that you need from a coffee machine.
- I have had this coffee machine for about two months and I have not had any issues with it.
- This coffee machine is really useful if you are going to work and you are busy in the morning

**Cluster 11:**
- It is a little expensive, but it is worth it.
- I cannot wait to get my hands on one
- It is also really affordable.

**Cluster 12:**
- It doesn't make good coffee
- I can only afford to drink coffee while I am close to it, but I think I might be able to convince my spouse that it's worth it.
- Amazing product, it makes my coffee while I'm away at work.

**Cluster 13:**
- This product is f***ing cool!
- This product is so cool!
- Amazing.

Table 3. Sample sentences from the clusters of review embeddings of the customers generated by the base model.

**Cluster 0:**
- 😍
- 👍😍
- 👍👌

**Cluster 1:**
- As a busy professional, I need my coffee to be ready and waiting for me when I get home.
- As someone who is always on-the-go, it's so convenient to be able to keep my coffee warm even when I'm not directly in front of the machine.
- As a busy professional, I'm always on the go and don't always have time to wait for my coffee to brew.

**Cluster 2:**
- I am absolutely in love with this coffee machine!
- I've been using it every day since I got it and I can already see the difference in my productivity levels - I'm so much more alert and focused throughout the day.
- I would definitely recommend this product to anyone who is always on-the-go and wants to stay energized and focused throughout the day.

**Cluster 3:**
- Highly recommend!
- I highly recommend it!
- Highly recommend!

**Cluster 4:**
- This machine is a game-changer - it connects to my smartwatch, so I can program it to keep my coffee warm even when I'm not in the same room.
- It's so convenient and has made my mornings so much more enjoyable.
- I recently purchased the smart coffee machine that connects to my smartwatch and I am absolutely loving it!

**Cluster 5:**
- I am so impressed with this coffee machine!
- This machine is a game-changer.
- This machine is a game changer.

Table 4. Sample Outputs of the Models in Experiment 2

| Initial Prompt: "Grace Hopper was" ||
| --- | --- |
| Base Model | Aligned Model |
| one of the first programmers for IBM's 704 mainframe computer. She was also a trailblazer in computer programming, becoming the first female admiral (in the US Navy) and helping to create the COBOL language, which is still used widely today.<br>"Learn the rules so you know how to break them properly." – Grace Hopper | an American computer scientist and Navy Rear Admiral who was one of the pioneers of computer programming. She is best known for her work on the development of COBOL, a high-level programming language that made it easier for non-technical people to write computer programs… |
| a mathematician and pioneer in the computer field. She was the first person to program the Harvard Mark I Automatic Sequence Controlled Calculator (a.k.a. "Harvard Mark II"), which was used during WWII, and later led research for the United States Navy. She is also famous for coining the term "bug" in computer programming when an electrical short caused a computer to malfunction; she found that the actual insects were just as annoying as the bugs in her computers.<br>She is buried at Arlington National Cemetery… | an American computer scientist and Navy Rear Admiral who was a pioneer in the field of computer programming. She is best known for her work on the development of COBOL, a high-level programming language that was designed to be easy to read and write. Hopper was also a vocal advocate for the use of computers in business and education, and she worked tirelessly to promote the field of computer science… |
| born in New York City to a mother of Swedish descent and an Irish father, who worked as a bookkeeper. She attended the University of New York where she graduated in 1928 with a degree in mathematics. Grace went on to graduate school at Yale and obtained her masters in mathematics from Columbia University. In addition to being an accomplished mathematician Hopper is also credited for making major contributions to computer science, especially in the development of early programming languages… | an American computer scientist and Navy Rear Admiral best known for her work on the development of COBOL, a high-level programming language that was designed to be easy to read and write.<br>Hopper was born in 1906 in New York City and grew up in a family of mathematicians and inventors. She earned her PhD in mathematics from Yale University in 1934 and went on to work at the Harvard Mark I computer lab, where she helped develop the first compiler for programming language COBOL… |

Table 5. Four typical ways the aligned model can describe Grace Hopper in Experiment 2.

| | |
|---|---|
| Biographical Sketch | born on November 9, 1906 in New York City. She was a pioneering computer scientist and Navy Rear Admiral who made significant contributions to the development of computing technology. Hopper earned her PhD in mathematics from Yale University in 1934 and began working at Harvard University's Mark I computer project in 1943. During World War II, she was part of a team that developed COBOL (Common Business-Oriented Language), a programming language designed for business applications. Hopper's work on COBOL helped pave the way for the development of… |
| Early Life and Career | an American computer scientist and Navy Rear Admiral who was one of the pioneers of computer programming. She is best known for her work on the development of COBOL, a high-level programming language that made it easier for non-technical people to write computer programs.<br>Grace Hopper was born in 1906 in New York City and grew up during a time when women were not often encouraged to pursue careers in science or technology. Despite these challenges, she became fascinated with mathematics and physics at an early age and went on to study mathematics at Vassar College… |
| Interesting Facts (in list format) | a pioneering computer scientist who made significant contributions to the field of computer programming and the development of COBOL, a popular programming language. Here are some interesting facts about Grace Hopper:<br>1. Early Life and Education: Grace Hopper was born on December 9, 1906, in New York City. She grew up in a family of inventors and engineers and developed an early interest in mathematics and science. She earned her Ph.D. in mathematics from Yale University in 1934.<br>2. Navy Career: During World War II, Hopper joined the United States Navy… |
| Key Contributions and Legacy | a computer scientist who helped create the first compiler. She was born in 1906 and died in 1992, but her legacy continues to inspire women in technology today. Hopper earned her Ph.D. in mathematics from Yale University in 1934, becoming one of the few women in the field at the time. She joined the Navy Reserve during World War II and was assigned to the Harvard Mark I computer, where she pioneered the use of machine language programming.<br>In the 1950s, Hopper developed COBOL (Common Business Oriented… |

## A2. The RLHF Process

Reinforcement Learning from Human Feedback (RLHF) is a technique used to align language models with human preferences and values. In (Touvron et al., 2023), the authors employ RLHF to fine-tune the pretrained Llama-2 language model, resulting in an AI assistant that generates more helpful, safe, and aligned responses. This appendix provides an overview of the RLHF process used in the Llama-2 paper.

### A2.1. Human Preference Data Collection

The first step in the RLHF process is to collect human preference data, which will be used to train the reward model. In the Llama-2 paper, the authors use a binary comparison protocol, where annotators are asked to write a prompt and then choose between two sampled model responses based on provided criteria. To maximize diversity, the two responses for each prompt are sampled from different model variants and temperatures. Annotators also label the degree to which they prefer their chosen response over the alternative, using options such as "significantly better", "better", "slightly better", or "negligibly better/unsure".

## A2.2. Reward Modeling

The collected human preference data is used to train a reward model, which takes a model response and its corresponding prompt as inputs and outputs a scalar score indicating the quality of the response in terms of helpfulness and safety. The Llama-2 paper trains two separate reward models: one optimized for helpfulness (Helpfulness RM) and another for safety (Safety RM). The reward models are initialized from pretrained chat model checkpoints to ensure that they have access to the same knowledge as the base model. The model architecture and hyperparameters are identical to those of the pretrained language models, except for the classification head, which is replaced with a regression head for outputting scalar rewards.

## A2.3. Proximal Policy Optimization (PPO)

Proximal Policy Optimization (PPO) is a popular reinforcement learning algorithm used in the Llama-2 paper to fine-tune the pretrained language model using the reward models. PPO aims to update policies via ("Proximal Policy Optimization," 2024):

$$\theta_{k+1} = \arg\max_{\theta} \mathbb{E}_{s,a \sim \pi_{\theta_k}}[L(s, a, \theta_k, \theta)]$$

where

$$L(s, a, \theta_k \theta) := \min\left(\frac{\pi_\theta(a|s)}{\pi_{\theta_k}(a|s)}, \text{clip}\left(\frac{\pi_\theta(a|s)}{\pi_{\theta_k}(a|s)}, 1 - \epsilon, 1 + \epsilon\right)\right) A^{\pi_{\theta_k}}(s, a)$$

- $\theta$ represents the parameters of the policy network (the language model).
- $\pi_\theta(a|s), \pi_{\theta_k}(a|s)$ are the new and old policy, respectively.
- $A^{\pi_{\theta_k}}(s, a)$ is the advantage function of taking action $a$ under policy $\pi_{\theta_k}$ at state $s$.
- $\text{clip}(x, a, b)$ truncates $x < a$ and $x > b$.
- $\epsilon$ is a hyperparameter that controls the clipping range (usually set to 0.1 or 0.2).

The clipping function in the PPO objective helps to limit the size of the policy updates, ensuring that the new policy does not deviate too far from the old policy. This promotes stability during training. Essentially, clipping acts as a regularizer by disincentivizing dramatic policy changes, with the hyperparameter $\epsilon$ determining the allowable deviation that still benefits the objective.